\documentclass[journal]{IEEEtran}

\usepackage{graphicx}
\usepackage{cite}
\usepackage{amssymb}
\usepackage{latexsym,amscd}

\usepackage{graphicx}
\usepackage{float}

\usepackage{times}

\usepackage{soul}
\usepackage{url}
\usepackage[hidelinks]{hyperref}
\usepackage[utf8]{inputenc}
\usepackage[small]{caption}
\usepackage{graphicx}
\usepackage{amsmath}
\usepackage{booktabs}
\usepackage{threeparttable}  
\usepackage{xcolor}

\usepackage{subfigure} 

\usepackage{algorithm}
\usepackage{algorithmic}

\usepackage{ifthen}
\usepackage{float}
\usepackage{amssymb}

\DeclareMathOperator*{\argmax}{arg\,max}

\DeclareMathOperator{\tr}{tr}
\ifCLASSINFOpdf
\else
\fi

\begin{document}
\title{$R^{2}$-Trans: Fine-Grained Visual Categorization with Redundancy Reduction}

\author{Yu~Wang$^{*}$,
        Shuo~Ye$^{*}$,
 		 Shujian Yu
        and~Xinge~You,~\IEEEmembership{Senior~Member,~IEEE}% <-this % stops a space

\thanks{This work was supported in part by the National Natural Science Foundation of China under Grant 11671161, Grant 61571205, and Grant 61772220, in part by the Key Program for International S\&T Cooperation Projects of China under Grant 2016YFE0121200, in part by the Special Projects for Technology Innovation of Hubei Province under Grant 2018ACA135, in part by the Key Science and Technology Innovation Program of Hubei Province under Grant 2017AAA017, in part by the Natural Science Foundation of Hubei Province under Grant 2018CFB691, and in part by the Science, Technology and Innovation Commission of Shenzhen Municipality under Grant  JCYJ20180305180637611, Grant JCYJ20180305180804836, and Grant JSGG20180507182030600. (Corresponding author: Xinge~You.)}
\thanks{Yu~Wang and Shuo~Ye are with the School of Electronic Information and Communications, Huazhong University of Science and Technology, Wuhan 430074, China.}% <-this % stops a space

\thanks{Shujian Yu is with the Department of Electrical and Computer Engineering, University of Florida, Gainesville, FL 32611 USA.}

\thanks{Xinge~You is with the School of Electronic Information and Communications, Huazhong University of Science and Technology, Wuhan 430074, China, and also with the Shenzhen Research Institute, Huazhong University of Science and Technology, Shenzhen 518000, China.}

\thanks{($*$ These authors contributed equally to this work and should be considered co-first authors.)}
}

\markboth{Journal of \LaTeX\ Class Files,~Vol.~14, No.~8, August~2015}%
{Shell \MakeLowercase{\textit{et al.}}: Bare Demo of IEEEtran.cls for IEEE Journals}

\maketitle

\begin{abstract}
Fine-grained visual categorization (FGVC) aims to discriminate similar subcategories, whose main challenge is the large intraclass diversities and subtle inter-class differences. Existing FGVC methods usually select discriminant regions found by a trained model, which is prone to neglect other potential discriminant information. On the other hand, the massive interactions between the sequence of image patches in ViT make the resulting class-token contain lots of redundant information, which may also impacts FGVC performance. In this paper, we present a novel approach for FGVC, which can simultaneously make use of partial yet sufficient discriminative information in environmental cues and also compress the redundant information in class-token with respect to the target. Specifically, our model calculates the ratio of high-weight regions in a batch, adaptively adjusts the masking threshold and achieves moderate extraction of background information in the input space. Moreover, we also use the Information Bottleneck~(IB) approach to guide our network to learn a minimum sufficient representations in the feature space. Experimental results on three widely-used benchmark datasets verify that our approach can achieve outperforming performance than other state-of-the-art approaches and baseline models.
The code of our model is available at the anonymous Github page: https://anonymous.4open.science/r/R-2-Trans.
\end{abstract}


\IEEEpeerreviewmaketitle

\section{Introduction}

\IEEEPARstart{F}{ine}-grained visual recognition~(FGVC) aims to classify images into subordinate categories. Compared with the traditional category images, fine-grained images have subtle inter-class and large intra-class variations. Therefore, the key to solving this problem is understanding fine-grained visual differences that sufficiently discriminate between objects that are highly similar in overall appearance but differ in fine-grained features~\cite{wei2021fine}. At present, relevant research has been widely used in animal and plant research, traffic security, commodity retail and other fields~\cite{zheng2020fine, liu2019bidirectional, zhang2019part, zhang2017picking, zhao2017diversified}.

\begin{figure}[t]
  \begin{center}
  \includegraphics[width=0.45\textwidth]{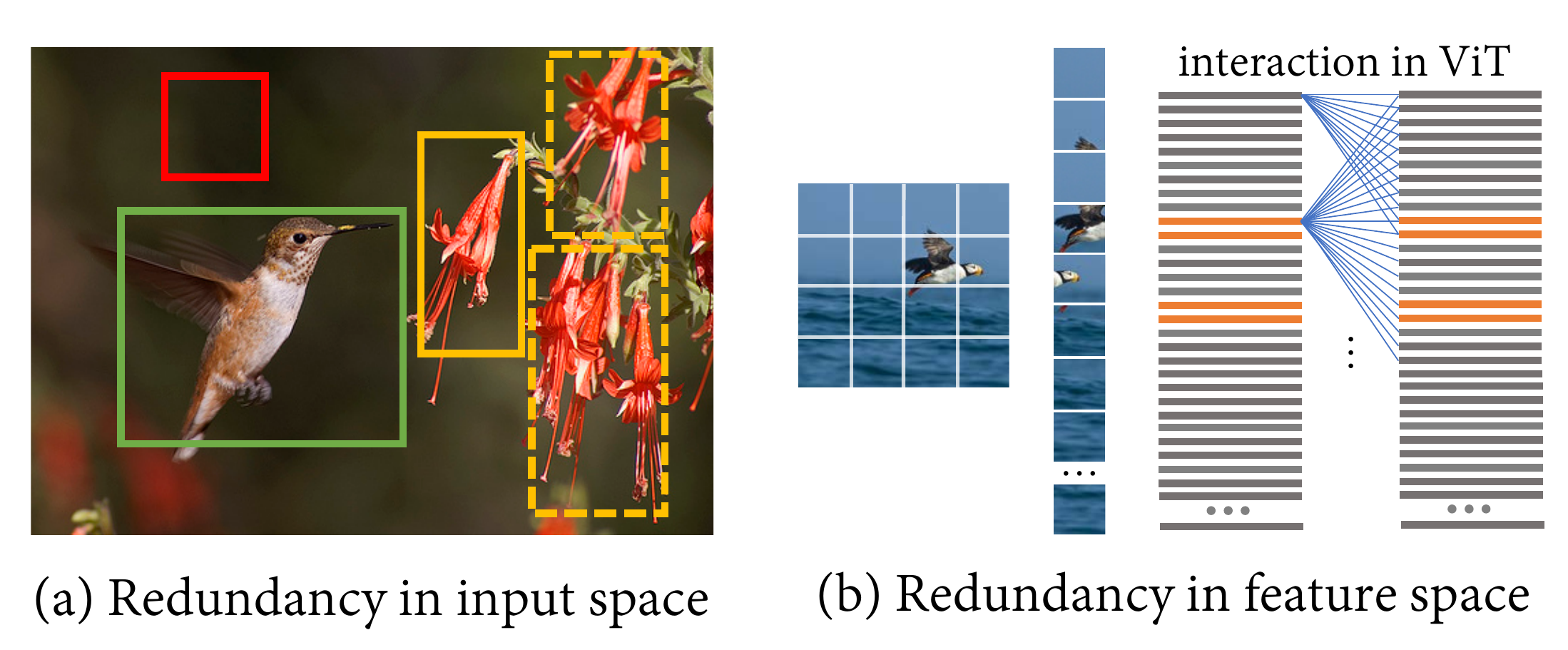}
  \end{center}
	\setlength{\abovecaptionskip}{-0.2cm} 
  \caption{Information redundancy in both (a) the input space; and (b) the feature space. The flowers ($a.k.a.$, the Hybrid fuchsia) are \emph{partially} informative to the target \textit{Hummingbird}, yet only one follow is sufficient to provide such discriminative information. On the other hand, the massive interactions between image patches in ViT may lead to it encodes the information from non-informative background areas as well. Our $R^2$-Trans aims to reduce the non-informative and redundant information in two spaces to facilitate FGVC performance.  
}
  \label{motivation}
\end{figure}

Earlier approaches~\cite{zhang2014part} make use of bounding box and part annotations to identify fine-grained regions or features. However, hand labeling is a labor-intensive and tedious job, and is prone to introduce human error ($a.k.a$., label noise). To address this issue, recent researches focused on recognition with only image-level labels. A series of models (e.g., \cite{zheng2017learning,fu2017look}) have been developed to use the effective information generated by itself for region localization and amplification. Among them, the convolutional neural network plays an important role by virtue of their translation invariance and local feature representation capability. To enhance the effect of target details in classification, attention mechanism is widely employed and has brought remarkable improvement. 

Recently, vision transformer~(ViT)~\cite{dosovitskiy2020image} has been applied to and achieved great success in traditional classification tasks owing to its multi-head self-attention mechanism for training. This enables the model to have a global view at an early stage and can fully interact with the features of different patches.   However, this overwhelming advantage did not extend to FGVC, one reason considering the fixed size of its patches, it makes ViT unable to generate multi-scale features to learn discriminative region attention.  
To handle this problem, the latest swin-transformer introduces the sliding window mechanism to enable the model to learn cross-patch information. CAMF~\cite{miao2021complemental} proposes a complemental attention multi-feature fusion network to extract multiple attention features to obtain more effective features. MST~\cite{li2021mst} dynamically masks some tokens of local patches without damaging the important foreground regions. Then, a global image decoder is used to reconstruct the original image based on the masked and unmasked tokens, forcing the model to learn spatial information.  
Since not every token has a target in it, one straightforward way to alleviate this problem is excluded immaterial token input in the final transformer layer, such as TransFG~\cite{he2021transfg} and FFVT~\cite{wang2021feature}. However, this kind of hard attention filtering is easy to fail, once the input image is small in resolution, filtered tokens may contain details which is difficult to obtain. Besides, both TransFG~\cite{he2021transfg} and FFVT~\cite{wang2021feature} cannot use all transformer layers attention information completely. For this reason, RAMS-Trans~\cite{hu2021rams} and AF-Trans~\cite{zhang2021free} were proposed successively. The latter case noticed that attention weights in each layer do not play an equal role in terminating the fused attention map, so, a selective attention collection module is designed, which gives different weights when integrating information between layers.  
Should be noted, previous methods are emphasized how to increase the repeated use of local area, however, general classification task will depend on the contour, color and other information, but the fine-grained targets often have similar outlines and colors, and uniform local amplification or enhancement will introduce redundant information and unnecessary noise. 
  
In this work, we systematically analyze the the bottleneck that restricts the performance of FGVC. In our perspective, the existence of redundant information in both input and feature spaces is a key issue. As shown in Fig.~\ref{motivation}(a), although the \textit{Hummingbird} in the green box is the target we want to identify, the surrounding environment is still \emph{partial} informative to the target. One can infer from the flowers ($a.k.a.$, the Hybrid fuchsia) in the yellow boxes that the target is not likely to be a \textit{Puffin} or a \textit{Seagull}. However, only one flower is sufficient to provide such discriminative information, the remaining two in the dashed boxes are relatively redundant. On the other hand, most of areas (such as the red box) in the background are non-informative and should be dropped out in the training. 
Similarly, there are also redundant information in the feature space. Such redundancy mainly comes from two aspects. First, similar contours and postures. General classification task will depend on the this information, but the fine-grained targets often have similar contours and postures. When the local features are amplified, some less important features are also enhanced.  
Second, as shown in Fig.~\ref{motivation}(b), the massive interactions between the sequence of image patches in ViT make the resulting class-tokens contain lots of redundant information, which may also encode information from non-informative areas (e.g., red box in Fig.~\ref{motivation}(a)).

For reasons above, we propose a novel approach for FGVC termed redundancy reduction transformer~($R^{2}$-Trans), which can simultaneously make use of partial yet sufficient discriminative information in environmental cues and also compress the redundant information in class-token with respect to the target. Specifically, our model consists of a batch-based dynamic mask module~(BDMM) and an information Bottleneck~(IB)~\cite{tishby99information} based loss function. Among them, BDMM is used to dynamically mask the target background in the input space, which can effectively reduce the redundancy information of the background. IB loss aims to guide our network to learn a minimum sufficient representations in the feature space. Compared to previous transformer based methods, we have several contributions: 

\begin{enumerate}
\item We systematically analyze the bottleneck that restricts the performance of FGVC and argue that a key issue is the large amount of redundancy in both input and feature spaces, especially when we use ViT as a backbone.
\item A batch-based dynamic mask method is proposed in the input space, which is able to identify the main target and the \emph{partial} informative discriminative regions in environmental cues with minimum redundancy.
\item We suggest compressing latent representations with IB approach, and introduce the recently proposed matrix-based R{\'e}nyi's $\alpha$-order entropy functional~\cite{giraldo2014measures} to simplify the implementation.
\item We conduct comprehensive evaluations on three widely used benchmark datasets. Experimental results demonstrate the superiority of our $R^{2}$-Trans over other 15 state-of-the-art (SOTA) approaches. 
\end{enumerate}

The rest of this paper is organized as follows. In Section II, we will introduce the related work including weakly supervised fine-grained visual recognition and basic theory of information bottleneck. We will describe our method in Section III. Experimental results will be presented in Section IV. Finally, Section V will draw a conclusion.

\section{Related Work}
In this section, we first present the work of weakly supervised FGVC. Then the most relevant work on information bottlenecks is briefly reviewed. 

\subsection{Fine-Grained Visual Categorization}   
In the past decade, the mainstream weakly supervised FGVC approaches can be categorized into two main paradigms including part localization based methods~\cite{zheng2017learning,fu2017look,yang2018learning} and end-to-end feature encoding methods~\cite{wang2017residual,cui2017kernel,luo2019cross}.  

Part localization based methods firstly employ subnetworks to initialize abundant region proposals and select the discriminative parts based on a specific strategy. Then, these selected areas are fed into new branches or sent back to the network again for learning. However, since the input is a part feature, this approach has difficulties in modeling the complicated relationship between the different distinct parts. Besides, multi-branch structure design leads to the complexity of network structure and makes optimize difficult. In addition, partial amplification of the target loses the potential discriminant information may contain in the background.  

Feature encoding methods circumvent this problem, they localize the corresponding high areas related to the image label and learn from it. The most typical approach is to encode higher-order statistics of convolutional features and extract compact holistic representations~\cite{lin2017bilinear,cui2017kernel}. However, those methods involves high feature dimension, which leads to poor interpretability. 

Recently, ViT~\cite{dosovitskiy2020image} based model demonstrate great potential in classification task, which employed its innate attention mechanism to capture the important regions in images~\cite{he2021transfg,wang2021feature}.  The most basic ViT flattens the segmented image patches and transforms them into patch tokens, then, the self-attention mechanism is used to link each Patch token to the Class token. This architecture forces the dissemination of information between patch tokens and class tokens, and link strength can be seen as an intuitive indicator of token importance. However, using ViT directly in FGVC does not take full advantage, because not every token has a target in it and this approach will leads to feature redundancy. One important reason considering its interaction mechanism, all tokens have interacted during the feed-forward process in multi-head attention layers~\cite{he2021transfg}, but not every token has a target in it. Besides, for fine-grained images, not all regions in target contribute uniformly to identification, this also intensifies the redundancy of features.  

Different from the above methods, our method can make use of partial yet sufficient discriminative information in environmental cues, which can effectively reduce the interference of non-informative or redundant regions in input space. Besides, we use the IB approach to force our $R^{2}$-Trans to learn a minimum sufficient representation to further reduce redundancy in feature space.

\subsection{Information Bottleneck} 

There is a recent trend to improve the understanding and the practical performances of deep neural networks using ideas or principles from Information Theory. 

Among different information-theoretic learning principles, the most notable one is the IB approach~\cite{tishby99information,tishby2015deep}, which can also be interpreted as an approximation to the minimum sufficient statistics~\cite{shamir2010learning,achille2018emergence}. Given input variable $X$ and desired response $Y$ (e.g., class labels), the IB approach aims to extract from $X$ a compressed representation $T$ that is most informative to predict $Y$. 
Formally, this objective is formulated as finding $T$ such that the mutual information $I(Y;T)$ is maximized, while keeping mutual information $I(X;T)$ below a threshold $\alpha$: 
\begin{equation}\label{eq:IB_original}\small
    \argmax_{T\in\Delta} I(Y;T)\quad \text{s.t.} \quad I(X;T)\leq \alpha,
\end{equation}
where $\Delta$ is the set of random variables $T$ that obey the Markov chain $Y - X - T$ (i.e., any information that $T$ has about $Y$ is from $X$). In practice, one is hard to solve the above constrained optimization problem of Eq.~(\ref{eq:IB_original}), and $T$ is found by maximizing the following IB Lagrangian:
\begin{equation}\label{eq:IB_Lagrangian}
  \mathcal{L}_{\text{IB}} = I(Y;T)-\beta I(X;T),
\end{equation}
where $\beta$ is a Lagrange multiplier that controls the trade-off between the performance of $T$ on task $Y$ (as quantified by $I(Y;T)$) and the complexity of $T$ (as measured by $I(X;T)$).

The IB approach has recently been applied to learn a compressed yet meaningful representations for lots of down-stream tasks, ranging from computer vision~\cite{zhmoginov2020information}, reinforcement learning~\cite{kim2021drop}, natural language processing~\cite{bang2021explaining}. To the best of our knowledge, there is only one work \cite{lai2021information} that has investigated IB approach on advanced visual recognition tasks including FGVC. Different from ours, \cite{lai2021information} developed an IB-inspired spatial attention module as a replacement of the traditional attention mechanism in CNNs. However, it is hard to extend this new module to ViT and the performance improvement on FGVC datasets is marginal compared to state-of-the-art (SOTA) CNN-based methods like \cite{ding2019selective}.  

\section{The Proposed Method}
\begin{figure*}[htbp]
  \begin{center}
  \includegraphics[width=0.99\textwidth]{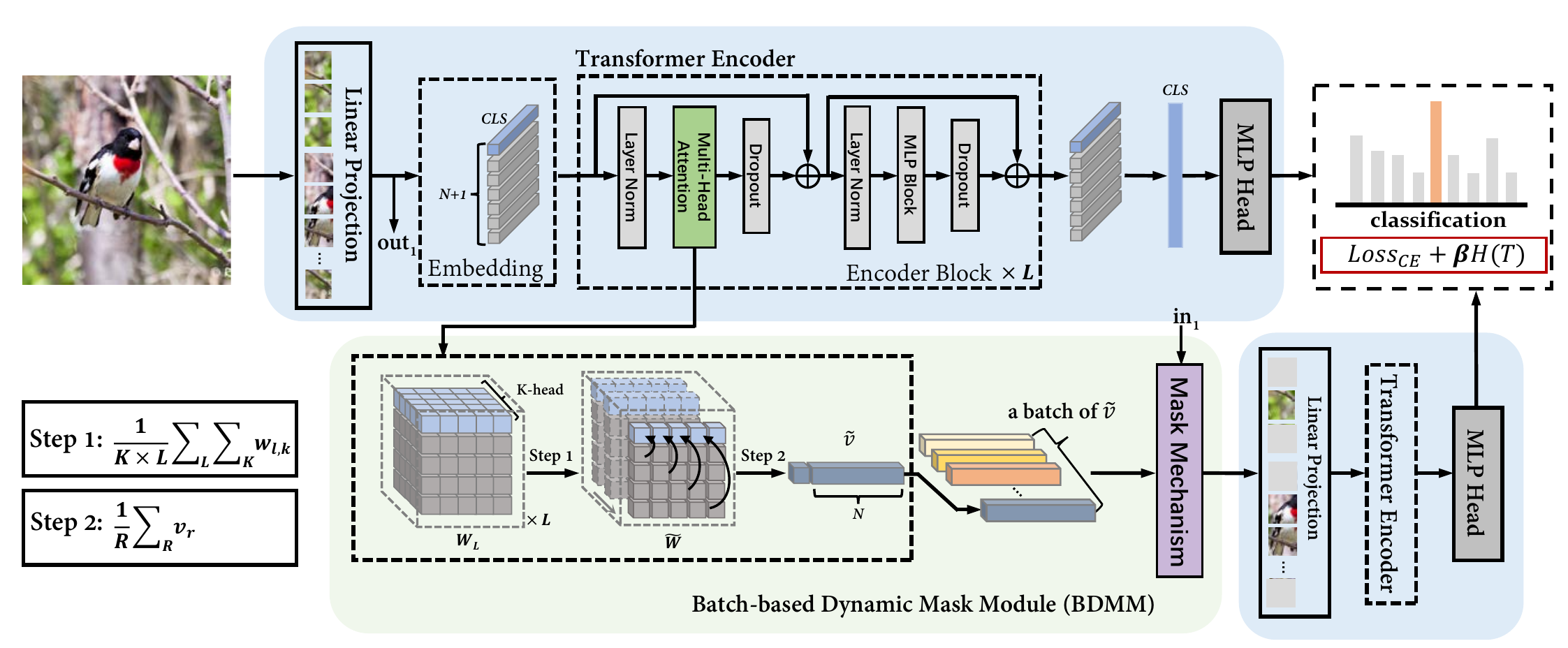}
  \end{center}
	\setlength{\abovecaptionskip}{-0.2cm} 
  \caption{Illustration of our proposed $R^{2}$-Trans framework. It is consists of a batch-based dynamic mask module~(BDMM) and an information bottleneck~(IB) based loss function. BDMM is used to reduce the redundancy information of the background in the input space. IB loss aims to guide the network to learn a minimum sufficient representation in the feature space.}
  \label{fig2}
\end{figure*}

In this section, we describe our $R^{2}$-Trans. As shown in Figure~\ref{fig2}, the proposed method consists of two core parts: batch-based dynamic mask module and information bottleneck loss. These two parts will be described thoroughly in the following subsections.

% which realizes the reduction of redundant features from input space and feature space. 

\subsection{Batch-based Dynamic Mask in Input Space}
As mentioned earlier, when ViT is used for FGVC, it inevitably brings a lot of noise and redundant information in the input space. Inspired by MST~\cite{li2021mst}, we design a BDMM to solve this problem. Specifically, the weight maps obtained by the multi-head attention module is used to measure the importance of different patches. Then, dynamically adjust different ratio in the unit of a batch to mask those non-informative and redundant patches, so as to ensure our network to learn a minimum sufficient representations. We follow the settings of ViT and use it as the backbone. We first reshape the input image $x \in \mathbb{R}^{H \times W \times C}$ into a sequence of flattened 2D patches $\mathbf{x}_{p} \in \mathbb{R}^{N \times\left(P^{2} \times C\right)}$, where $(H, W)$ is the size of the raw image, $C$ is the number of channels, $(P, P)$ is the size of each patch, and $N = H\times W/P^{2}$ is the resulting number of patches. Then we flatten the patches and map to $D$ dimensions with a trainable linear projection to obtain the patch tokens $x_{patch}\in \mathbb{R}^{N\times D}$. A learnable class token $x_{cls} \in \mathbb{R}^{1 \times D}$ is embedded into the patches to get $N+1$ tokens, which can be formulated as:
\begin{equation}
    \boldsymbol{x_{0}}=[x_{cls}||x_{patch}]+x_{pos},
\end{equation}
where $x_{0} \in \mathbb{R}^{(N+1) \times D}$ is the input of the first transformer block, and $x_{pos} \in \mathbb{R}^{(N+1) \times D}$ is the learnable position embedding which retains positional information.

The transformer encoder is composed of a stack of L transformer blocks, which consists of multi-head self-attention~(MSA) and MLP blocks. We can get the weight maps from the MSA module,
which can be expressed as:
\begin{equation}
    \boldsymbol{W_{L}}=[w_{1}, w_{2}, ... ,w_{L}]
\end{equation}
\begin{equation}
    \boldsymbol{W_{l}}=[w_{l,1}, w_{l,2}, ... ,w_{l,K}]
\end{equation}
here $L$ is the number of transformer blocks, $K$ is the number of attention headers of each block 
and $w_{l,k} \in \mathbb{R}^{({N+1}) \times ({N+1})}$ is one weight map.
We integrate the weight map of all heads at different blocks:

\begin{equation}
    \widetilde{\boldsymbol{W}} = \frac{1}{LK} \sum^{L}_{l=1} \sum^{K}_{k=1} w_{i,k},
 \end{equation}
where $\widetilde{\boldsymbol{W}} \in \mathbb{R}^{({N+1}) \times ({N+1})}$ is fully consider the influence of different levels of information. 

To make full use of the information contained in each patch and emphasize the importance of the object in the image, we take the average of the row vectors $v_{r} \in \mathbb{R}^{1 \times (N+1)}$ of $\tilde W_{l}$. Then, the fully fuse vector $\widetilde v \in \mathbb{R}^{1 \times (N+1)}$ is obtained. This specific process can be described as: 
\begin{equation}
    \widetilde v = \frac{1}{R} \sum^{R}_{r=1} v_{r},
\end{equation}
After removing the first class of related patch, we get the patch importance map $\widetilde v_p \in \mathbb{R}^{1 \times N}$, which comprehensively represents the importance of $N$ input patches. At last, $\widetilde v_p$ is used to guide which patch should be masked in $x_p$. Since the scale of the instances in each image are different, we take batch as the unit to generate the mask ratio. The algorithm for getting the masking ratio of the current batch is as follows:

\begin{algorithm}[h]
	\caption{Get Masking Ratio of Current Batch}
	{\bf Input:} A batch patch importance maps $v_{B}=[\widetilde v_p^1, \widetilde v_p^2, ... , \widetilde v_p^B]$, 
    an adjustment factor $\lambda (\lambda > 0)$ \\
	{\bf Require:} The ratio to mask for the current batch $r_{batch}$ \\
	\begin{algorithmic}[1]
        \STATE Create an empty list $r_{list}$
		\FOR {$\widetilde v_p^i$ in $v_{B}$}
		\STATE $m_{i}$ $\leftarrow$ Calculate the average value of current $\widetilde v_p^i$
		\STATE $r_{i}$ $\leftarrow$ Calculate the ratio of patch whose weight is less than $m_{i} \cdot \lambda$ in $\widetilde v_p^i$
		\STATE Append $r_{i}$ to $r_{list}$
		\ENDFOR		
    	\STATE $r_{batch}$ $\leftarrow$ Get average result of $r_{list}$
    	\STATE \bf Return $r_{batch}$
	\end{algorithmic}
\end{algorithm}

Since each patch token in the ViT is fully integrated with the other patch tokens after transformer encoder, the DBMM returns to the original input to mask these non-informative patches. Then, the useful patches are sent into the transformer encoder again to get the new class token ($x^{0}_{L}$), which is used with the old $x^{0}_{L}$ for training. During this process, ViT parameters are shared. Note that, in the test phase, we directly input the original image without BDMM step, so our approach is fast to predict in practical applications.

\begin{figure*}[htbp]
\centering
\subfigcapskip=5pt
\begin{tabular}{ccc}
\subfigure[CUB-200-2011]
{\includegraphics[width=0.32\linewidth]{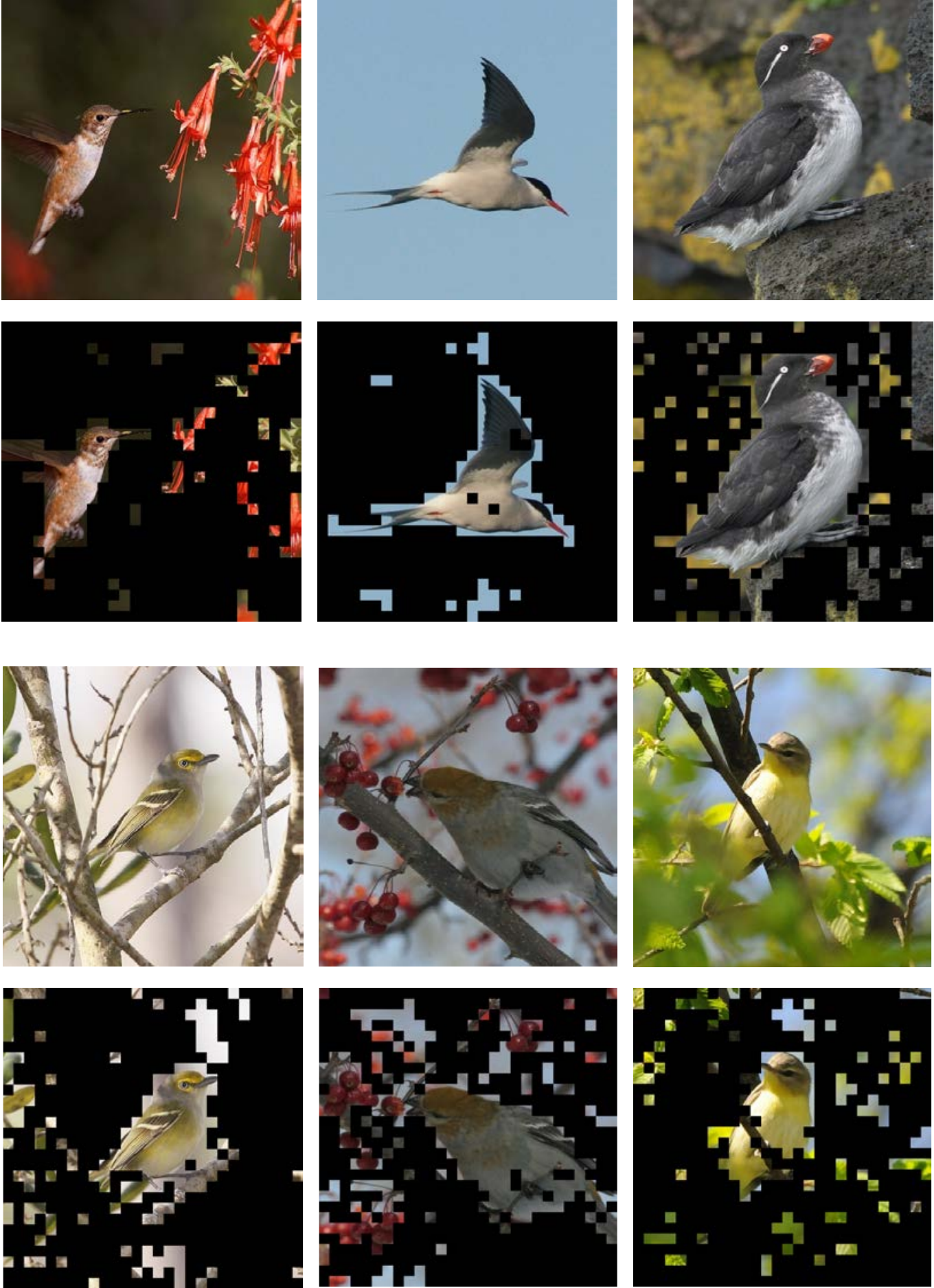}}\hspace{-2mm}
& 
\subfigure[Stanford Dogs]
{\includegraphics[width=0.32 \linewidth]{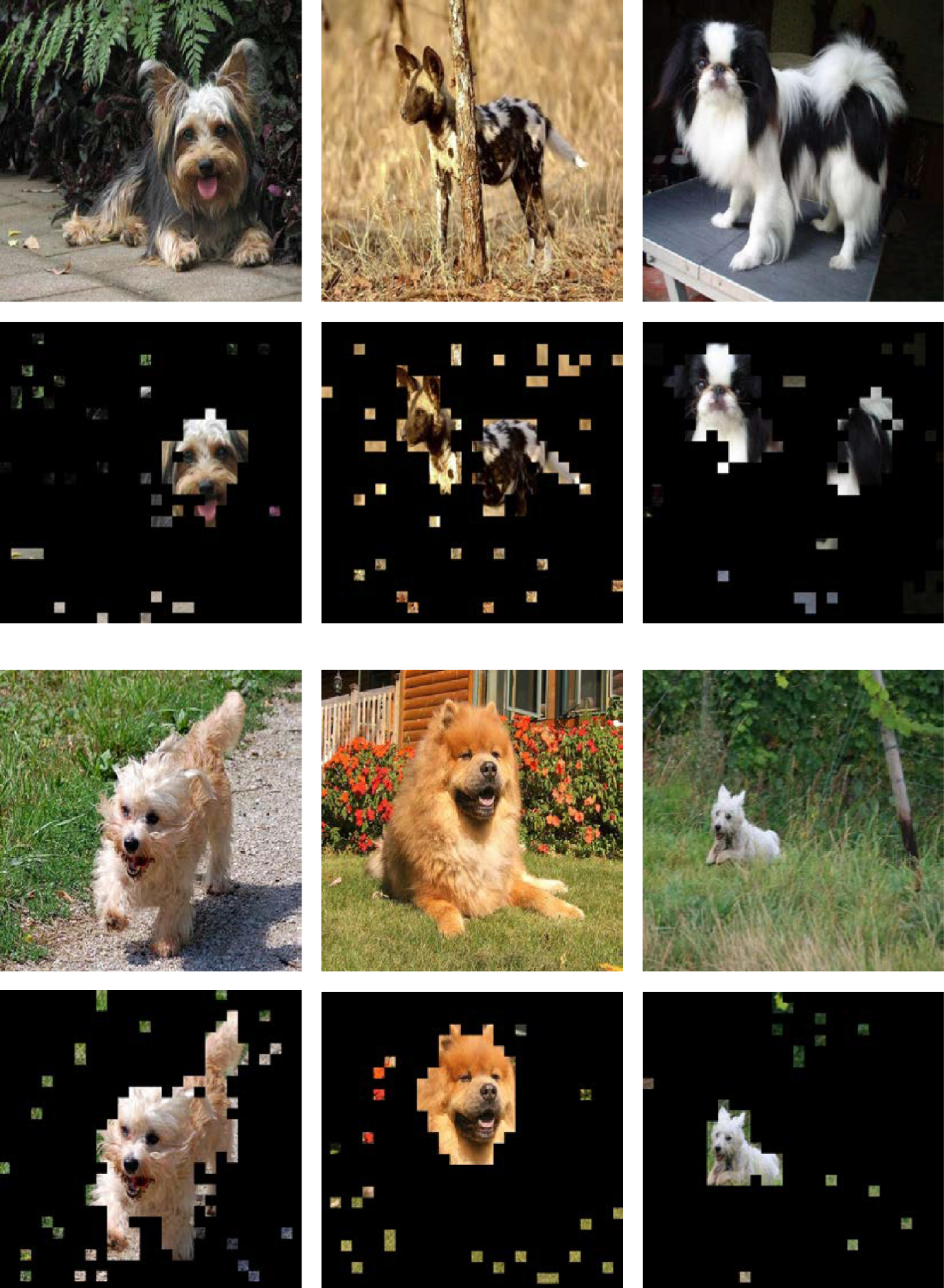}}\hspace{-2mm}
& 
\subfigure[NABirds]
{\includegraphics[width=0.32 \linewidth]{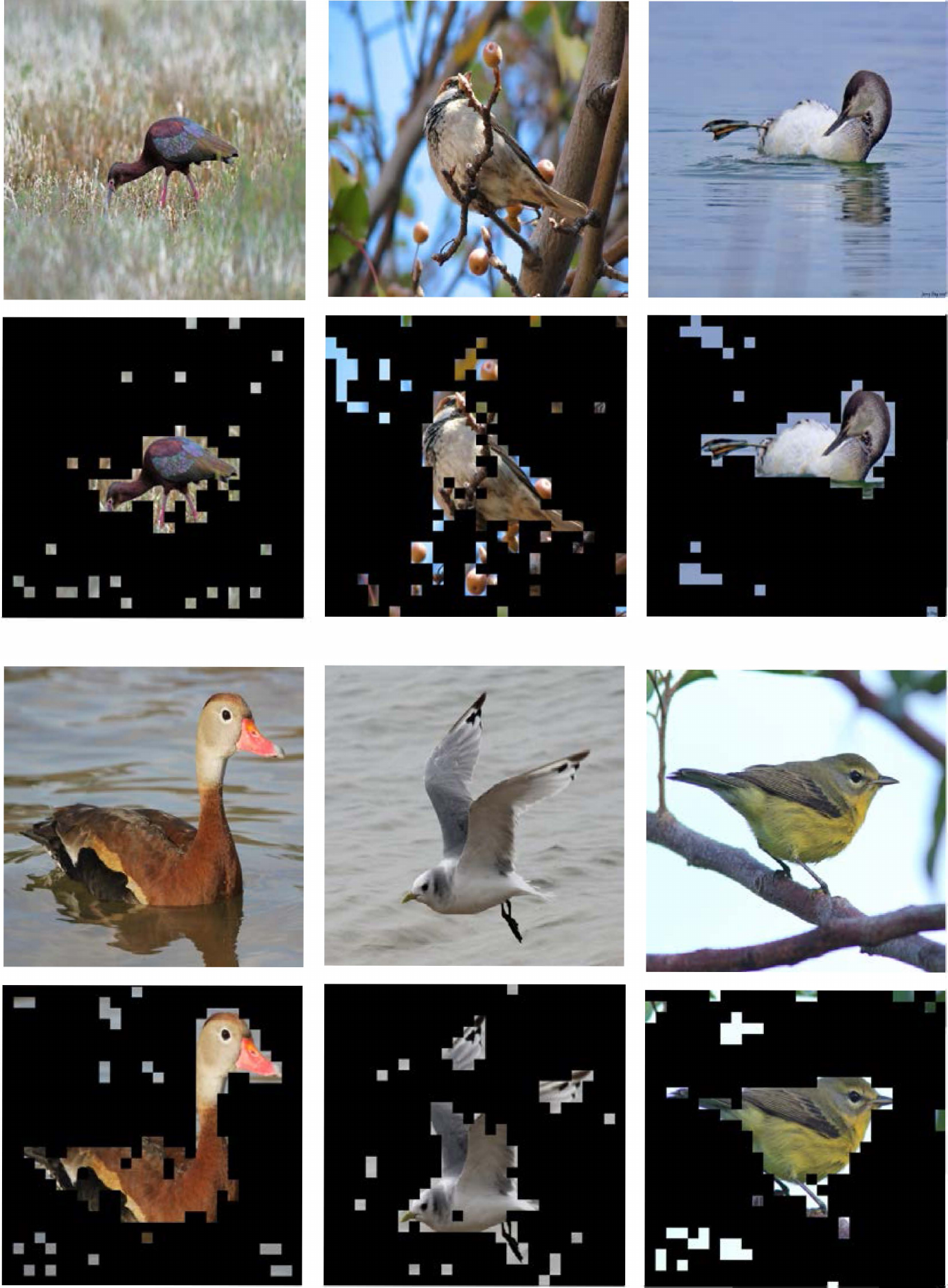}}
\end{tabular}
\setlength{\abovecaptionskip}{-0.1cm} 
\caption{Visualization results of BDMM on CUB-200-2011 dataset. The first line is the original image, and the second line is the original image after dynamic mask.}
\label{BDMM}
\end{figure*}

We demonstrate its effect in Figure~\ref{BDMM}, it can be seen that the generated mask covers most of the non-informative patches, effectively reducing the redundancy in the input space. However, we can still distinguish the target environment, which shows that the environment does provide potential discriminating information, which is consistent with our intuition.

\subsection{Information Bottleneck in Feature Space}
%Fine-grained visual recognition completes the classification of subclasses through slight detail differences. Noise or redundant information may affect the recognition. In the past, the re-location method was used to solve this problem, but it made the model structure complex and optimization difficult. 

We use IB approach to remove the redundancy in feature space by the objective of $\max I(Y;T)-\beta I(X;T)$, in which $T$ refers to the class tokens extracted after Transformer encoder. 
There are different ways to parameterize IB by neural networks. In general, the maximization of $I(Y;T)$ is equivalent to the minimization of cross-entropy (CE) loss~\cite{alemi2017deep,amjad2019learning}, which turns the objective of deep IB into a standard CE loss regularized by a differentiable mutual information term $I(X;T)$.

Most of existing deep IB approaches differ in the way to estimate $I(X;T)$. Depends on implementation details, $I(X;T)$ can be measured by variational approximation~\cite{alemi2017deep,kolchinsky2019nonlinear}, mutual information neural estimator (MINE)~\cite{belghazi2018mutual} and the matrix-based R{\'e}nyi's $\alpha$-order entropy functional~\cite{yu2021deep}.

%In practice, . Therefore, Eq.~(\ref{eq:IB_Lagrangian}) can be simply realized by a cross entropy loss combined with a regularization on $\mathbf{I}(X;T)$. 

In this work, same to ~\cite{yu2021deep}, we also use the matrix-based R{\'e}nyi's $\alpha$-order entropy functional to estimate information-theoretic quantities directed on the reproducing kernel Hilbert space~(RKHS) formed by the projected samples, which avoids variational approximation and the tuning of an additional neural network (as in MINE). However, instead of estimating $I(X;T)$, we straightforwardly regulate the entropy of features (i.e., $H(T)$) for simplicity. Thus, our final objective reduces to:
\begin{equation}
  Loss_{\text{IB}}= Loss_{\text{CE}}+\beta H(T).
\end{equation}
We choose $\beta=0.005$ throughout this work. 
Note that, regulating $H(T)$ as an alternative to $I(X;T)$ has also been recently used in \cite{ahuja2021invariance}. Mathematically, for a deterministic and feed-forward neural network, we have $H(T)$ = $I(X;T)$~\cite{amjad2019learning}~\cite{saxe2019information}.  

We finally present details on estimating $H(T)$ in a mini-batch for completeness. Specifically, given $N$ pairs of samples from a mini-batch of size $N$, i.e., $\{\mathbf{x}^m,\mathbf{t}^m\}_{m=1}^{N}$, each $\mathbf{x}^m$ be an input sample and each $\mathbf{t}^m$ denotes its class token extracted by Transformer encoder $\phi$. We can view both $\mathbf{x}$ and $\mathbf{t}$ as random vectors.
According to~\cite{giraldo2014measures}, the entropy of $\mathbf{t}$ can be defined over the eigenspectrum of a (normalized) Gram matrix $K_{\mathbf{t}}\in \mathbb{R}^{N\times N}$ ($K_{\mathbf{t}}(m,n)=\kappa(\mathbf{t}^{m}, \mathbf{t}^{n})$ and $\kappa$ is a Gaussian kernel) as:
\begin{equation}\label{eq:Renyi_entropy}\small
H_{\alpha}(A_\mathbf{t})=\frac{1}{1-\alpha}\log_2 \left(\tr (A_{\mathbf{t}}^{\alpha})\right)=\frac{1}{1-\alpha}\log_{2}\left(\sum_{m=1}^{N}\lambda _{m}(A_\mathbf{t})^{\alpha}\right),
\end{equation}
where $\alpha>0$. $A_\mathbf{t}$ is the normalized version of $K_\mathbf{t}$, i.e., $A_\mathbf{t}=K_\mathbf{t}/\tr(K_\mathbf{t})$. $\lambda_{m}(A_\mathbf{t})$ denotes the $m$-th eigenvalue of $A_\mathbf{t}$. Same to ~\cite{yu2021deep}, we choose $\alpha=1.01$.

Our IB regulariziation in the feature space can significantly reduce feature redundancy to seek a minimum yet sufficient representation, as will be illustrated in Section~\ref{sec:ablation}.

\section{Experiments}
To evaluate the proposed method, comprehensive experiments are conducted in the paper. The section includes four subsections. In section A, a brief introduction about the datasets and training strategy for the proposed model is presented. In section B, the experiment setting is introduced. In section C, the performance of the proposed method is compared with state-of-the-art methods. In section D, an ablation study is designed to demonstrate the effect of each module.

\subsection{Datasets and Implementation Details}
Experiments are conducted on three widely used datasets, namely CUB-200-2011~\cite{wah2011caltech}, Stanford Dogs~\cite{khosla2011novel}, and NABirds~\cite{van2015building}. All datasets provide fixed train and test splits.
The brief introduction for them is as follows:

{\bf CUB-200-2011}~\cite{wah2011caltech} is an extended version of the CUB-200 dataset. It includes 11,788 images from 200 different bird species and there are about 60 images for each species . Each image in this dataset is associated with detailed annotations including image-level labels, object bounding boxes, part locations and binary attributes. Only the images and their labels are utilized. 

{\bf Stanford Dog}~\cite{khosla2011novel} consists of 20580 images from 120 classes of dogs, which are divided into 12000 images for training and 8580 images for testing. Only the images and their labels are utilized. 

{\bf NABirds}\cite{van2015building} is a larger fine-grained dataset than the CUB-200-2011 dataset. It contains 555 classes with a total of 48562 North American bird images, of which 23,929 are used for training and the rest for testing. Should be note, many methods with complex operations or are not easy to experiment on a data set of this order of magnitude.

%The detailed statistics are summarized in Table~\ref{tab:tab1}. 
%%about these three datasets including class numbers and train/test distributions
%\begin{table}[h]
%  \setlength{\tabcolsep}{4pt}
%  \begin{center}
%	\setlength{\abovecaptionskip}{-0.04cm}
%	\caption{Summary statistics of fine-grained datasets.}
%	\label{tab:tab1}
%  \begin{tabular}{|c|c|c|c|}
%  \hline
%  Datasets & Category & Training & Testing \\
%  \hline\hline
%  CUB-200-2011                & 200 & 5994   & 5794 \\
%  Stanford Dogs               & 120 & 12000  & 8580 \\
%  NABirds                     & 555 & 23929  & 24633 \\
%%  iNat2017~\cite{van2018inaturalist}                 & 5089& 579184 & 95986 \\
%  \hline
%  \end{tabular}  \end{center}  
%  \vspace{-0.4cm} 
%\end{table} 

In our experiments, we use ViT-B-16 pre-trained on ImageNet21k as the backbone, and the input image size is $448\times448$. Our preprocessing follows the popular configurations~\cite{he2021transfg,wang2021feature}. Specifically, we use data augmentations, including random cropping and horizontal flipping during the training procedure. Only the center cropping is involved in inference. The model is trained with the stochastic gradient descent~(SGD) with a batch size of 24 and momentum of 0.9 for all datasets. The learning rate is set to 3e-2 in CUB and NABirds, and 3e-3 in Dogs. The schedule applies a cosine decay function to an optimizer step. The $\lambda$ is set to 1. Our implementation is based on PyTorch with a A6000 GPU. For all experiments, we adopt the top-1 accuracy as the evaluation metric and denotes it as "Acc."

\subsection{Model Configuration}
Model configuration experiments are conducted on the CUB-200-2011 dataset to verify the validity of the individual component and to determine the optimal hyperparameters. 

\textbf{Mask Ratio:}
To explore the influence of mask ratio and the effectiveness of our proposed BDMM in input space, a group of experiments of different mask ratios were conducted, and the results are shown in Table ~\ref{tab:tab2}. It can be seen that the mask ratio achieves the best performance between 0.5 and 0.7, which is caused by the compression of redundant features in the input space of the model. However, there is still a certain gap compared to our BDMM. This proves the validity of our algorithm.  

\renewcommand{\arraystretch}{1.2}
\begin{table}[h]
\setlength{\tabcolsep}{8pt}
\setlength{\abovecaptionskip}{-0.04cm}
\begin{center}
\caption{Experimental results using varied mask ratio.}
  \label{tab:tab2}
 \begin{tabular}{c|ccc cc|c}  
 \bottomrule 
  Ratio  & 0.1 & 0.3 & 0.5 & 0.7 & 0.9 & BDMM  \\
  \hline
  Acc. & 90.4  & 90.5 & 91.0 & 91.0 & 90.6 &\bf 91.3  \\
  \hline  
\end{tabular}
\end{center} 
\vspace{-0.1cm} 

\end{table}

\textbf{Information Compression ($\boldsymbol{\beta}$):}
To verify the effectiveness of loss function and investigate the effect of information compression $\beta$, we conduct extensive experiments on $\beta$ and the results are given in Table~\ref{tab:tab3}. As expected, information compression significantly enhances the network representation capability, which demonstrates that a large number of redundant features are learned by the model during the training process. Besides, we observe that increasing $\beta$ leads to higher accuracy. However, the performance slightly drops when the balance parameter $\beta$ increases from 0.005 to 0.01, which infers that when $\beta$ is equal to 0.005, the redundant information contained in the target has been squeezed out. 

\renewcommand{\arraystretch}{1.2}
\begin{table}[h]
\setlength{\tabcolsep}{8pt}
\setlength{\abovecaptionskip}{-0.04cm}
 \begin{center}
\caption{Experimental results using varied $\beta$.}
  \label{tab:tab3}
 \begin{tabular}{c|ccc ccc}  
 \bottomrule 
  $\beta$  & 0.0001 & 0.0005 & 0.001 & 0.005 & 0.01 & 0.05  \\
  \hline
  Acc. & 90.5  & 90.8 & 90.9 & \bf 91.2 & 90.8 & 90.6 \\
  \hline  
\end{tabular} \end{center}
\vspace{-0.4cm}  
\end{table}

\textbf{Batch Size:}
Since mask ratio is related to batch size, we also designed a set of experiments to illustrate this impact, and reported the results in Table~\ref{Batch}. 

\renewcommand{\arraystretch}{1.2}
\begin{table}[h]
  \setlength{\tabcolsep}{8pt}
  \begin{center}
	\setlength{\abovecaptionskip}{-0.04cm}
	\caption{Experimental results using varied batch size.}
	\label{Batch}
 \begin{tabular}{c|ccc cccc}  
 \bottomrule 
  Size  & 4 & 8 & 16 & 24 & 32 & 40 & 48\\
  \hline
  Acc. & 90.6  & 91.4 & 91.4 & \bf 91.5 & 91.0 & 91.1 & 91.0\\
  \hline  
\end{tabular} \end{center}
\vspace{-0.4cm}  

\end{table}

It can be seen that different batch sizes have a certain impact on our methods. As ViT tends to be larger batch, the optimal result does not appear when batch size=4. This situation is improved when the batch size increases. However, batch-based effect cannot be reflected when it is increased to a certain extent. When batch size =32, the result is close to fixed mask. Therefore, we use $\beta$=0.005 and batch size=24 in all the following experiments as it ideally balances the computational complexity and accuracy.

\subsection{Performance Evaluation}
%%%%%%%%%%%%%%%%%%%%%%%%%%%%%%%%%%%%%%%%%%--table for Classification Results--%%%%%%%%%%%%%%%%%%%%%%%%%%%%%%%%%%%%%%%%%%
This subsection presents the experimental results and analyses of our $R^{2}$-Trans approach compared with recent state-of-the-art methods on three widely used datasets. For a fair comparison, the backbone of the models is listed. To avoid the impact of different operating environments, when comparing the ViT as the backbone network model, we reproduced the open-source algorithm and repeated experiments in our environment according to the parameters of their paper, those repeated methods were marked with $*$.  All models are implemented under weakly supervised conditions and do not require additional annotation information.

\textbf{Comparison experiment on CUB-200-2011:}
We compare our methods with one up-to-date baseline model fourteen state-of-the-art methods. The results are reported in Table~\ref{Result1}. As can be seen, the accuracy of part localization methods such as MA-CNN and MA-CNN have a gap compared with our results. One reason considering those methods need to generate a large number of region proposals at first, but most of the regions inevitably contain environmental noise. DCL emphasizes local details by breaking global structures and then reconstructing images to learn semantic associations between local areas. However, the size of the segmented images remained the same throughout the training process, which meant that the model was difficult to deal with targets of different scales. It is worth noting that in all the CNN-based methods, S3N has the best performance, one reason for this result considering S3N exploits selective sampling to reduce the interference of background information. Unfortunately, it ignored redundant features in itself, which may prevent it from getting better results.

Compare with a model using the same backbone network, FFVT and AF-trans performed similarly to us. However, these approaches does not achieve the best performance due to insufficient consideration of redundancy in feature space. In general, in CUB-200-2011, compared with the model based on convolutional neural network, $R^{2}$-Trans achieves a 3.2\% improvement on top-1 accuracy and 1.2\% improvement compared to our base framework ViT. 

\renewcommand{\arraystretch}{1.2}
\begin{table}[h]
  \setlength{\tabcolsep}{8pt}
 \begin{center}
	\setlength{\abovecaptionskip}{-0.04cm}
 \caption{Acc. of our proposed method and several recent state-of-the-art methods on the CUB-200-2011 dataset.}
  \label{Result1} 
    \begin{tabular}{|c|c|c|}  
    \bottomrule   
	\textbf{Method}  & \textbf{Backbone}   & \textbf{CUB}   \\  
    \hline  \hline
HIHCA~\cite{cai2017higher}      & VGG-16	& 85.3     \\
MA-CNN~\cite{zheng2017learning} & VGG-19    & 86.5    \\
RA-CNN~\cite{fu2017look}        & VGG-19    & 85.3    \\
P-CNN~\cite{han2019p}           & VGG-19    & 87.3    \\
\hline
B-CNN~\cite{lin2017bilinear}    & ResNet-50   & 84.0  \cr  
DCL~\cite{wang2017residual}     & ResNet-50   & 86.2     \cr
Cross-X~\cite{luo2019cross}     & ResNet-50   & 87.7    \cr
MCL~\cite{chang2020devil}       & ResNet-50   & 87.3     \cr  
S3N~\cite{ding2019selective}    & ResNet-50   & 88.5     \cr 
MOMN~\cite{min2020multi}        & ResNet-50   & 88.3     \cr
\hline
ViT*~\cite{dosovitskiy2020image} & ViT-B-16    & 90.3    \cr   
TransFG*~\cite{he2021transfg}    & ViT-B-16    & 91.3    \cr	
%TPSKG~\cite{liu2021transformer}  & ViT-B-16    & 91.3   \cr  	
FFVT*~\cite{wang2021feature}     & ViT-B-16    & 91.5    \cr
RAMS-Trans~\cite{hu2021rams}     & ViT-B-16    & 91.3    \cr
AF-Trans~\cite{zhang2021free}    & ViT-B-16    & 91.5    \cr
$R^{2}$-Trans                    & ViT-B-16    & \bf 91.5   \cr
\hline
 \end{tabular}  \end{center}  
 \vspace{-0.4cm}
\end{table} 

\textbf{Comparison experiment on Stanford Dogs:}
We compare our methods with one up-to-date baseline model ten state-of-the-art methods. The results are reported in Table~\ref{Result2}.

\renewcommand{\arraystretch}{1.2}
\begin{table}[h]
  \setlength{\tabcolsep}{8pt}
 \begin{center}
	\setlength{\abovecaptionskip}{-0.04cm}
 \caption{Acc. of our proposed method and several recent state-of-the-art methods on the Stanford Dogs dataset.}
  \label{Result2} 
    \begin{tabular}{|c|c|c|}  
    \bottomrule   
	\textbf{Method}  & \textbf{Backbone}    & \textbf{Dogs}  \\  
    \hline  \hline
RA-CNN~\cite{fu2017look}        & VGG-19      & 87.3  \cr
P-CNN~\cite{han2019p}           & VGG-19      & 90.6  \\
\hline
SEF~\cite{luo2020learning}      & ResNet-50   & 88.8     \cr
Cross-X~\cite{luo2019cross}     & ResNet-50   & 88.9     \cr
MOMN~\cite{min2020multi}        & ResNet-50   &91.3      \cr
API-Net~\cite{zhuang2020learning} & ResNet-101   &90.3      \cr
\hline
ViT*~\cite{dosovitskiy2020image} & ViT-B-16     & 92.0   \cr   
TransFG*~\cite{he2021transfg}    & ViT-B-16     & 92.1   \cr	
%TPSKG~\cite{liu2021transformer}  & ViT-B-16    & 92.5   \cr  	
FFVT*~\cite{wang2021feature}     & ViT-B-16     & 91.4     \cr
RAMS-Trans~\cite{hu2021rams}     & ViT-B-16     & 92.4   \cr
AF-Trans~\cite{zhang2021free}    & ViT-B-16     & 91.6   \cr
$R^{2}$-Trans                    & ViT-B-16     & \bf 92.8   \cr
\hline
 \end{tabular}  \end{center}  
 \vspace{-0.4cm}
\end{table} 

In Stanford Dogs dataset, we also have the same observation as on the CUB-200-2011 dataset. From the results, we find that the proposed $R^{2}$-Trans outperforms all previous methods on Dogs dataset. RAMS-Trans has the closest performance to us, but it involve complex alternate optimization or multistage training strategies, which may cause information loss.

\textbf{Comparison experiment on NABirds:}
Similar experiments were performed on the NAbirds dataset to further validate the effectiveness and efficiency of our method. The results are reported in Table~\ref{Result3}.

\renewcommand{\arraystretch}{1.2}
\begin{table}[h]
  \setlength{\tabcolsep}{8pt}
 \begin{center}
	\setlength{\abovecaptionskip}{-0.04cm}
 \caption{Acc. of our proposed method and several recent state-of-the-art methods on the NABirds dataset.}
  \label{Result3} 
    \begin{tabular}{|c|c|c|}  
    \bottomrule   
	\textbf{Method}  & \textbf{Backbone}  & \textbf{NABirds}  \\  
    \hline  \hline

Cross-X~\cite{luo2019cross}     & ResNet-50    &86.2\cr
DSTL~\cite{cui2018large}			&Inception-v3  &87.9\cr
API-Net~\cite{zhuang2020learning}  & DenseNet-161 &88.1 \cr
FixSENet~\cite{touvron2019fixing}  & SENet-154    &89.2 \cr
\hline
ViT*~\cite{dosovitskiy2020image} & ViT-B-16    &89.6 \cr   
TransFG*~\cite{he2021transfg}    & ViT-B-16    &90.0 \cr	
%TPSKG~\cite{liu2021transformer}  & ViT-B-16   & 90.1 \cr  	
FFVT*~\cite{wang2021feature}     & ViT-B-16    & 89.7   \cr
$R^{2}$-Trans                    & ViT-B-16    & \bf 90.2 \cr
\hline
 \end{tabular}  \end{center}  
 \vspace{-0.4cm}  
\end{table} 

As can be seen, compared with the CNN-based methods, ViT-based methods achieved general improvement. One possible reason is that targets in the NABirds dataset vary greatly in scale, and large areas of environment introduce a lot of redundancy for identification. Our approach effectively suppresses this problem and achieves the best performance of the ViT-based approach.

\subsection{Ablation Study} \label{sec:ablation}

\begin{figure}[htbp]
\centering
\subfigure[CUB-200-2011]{\includegraphics[height=1.4in,width=3.4in,angle=0]{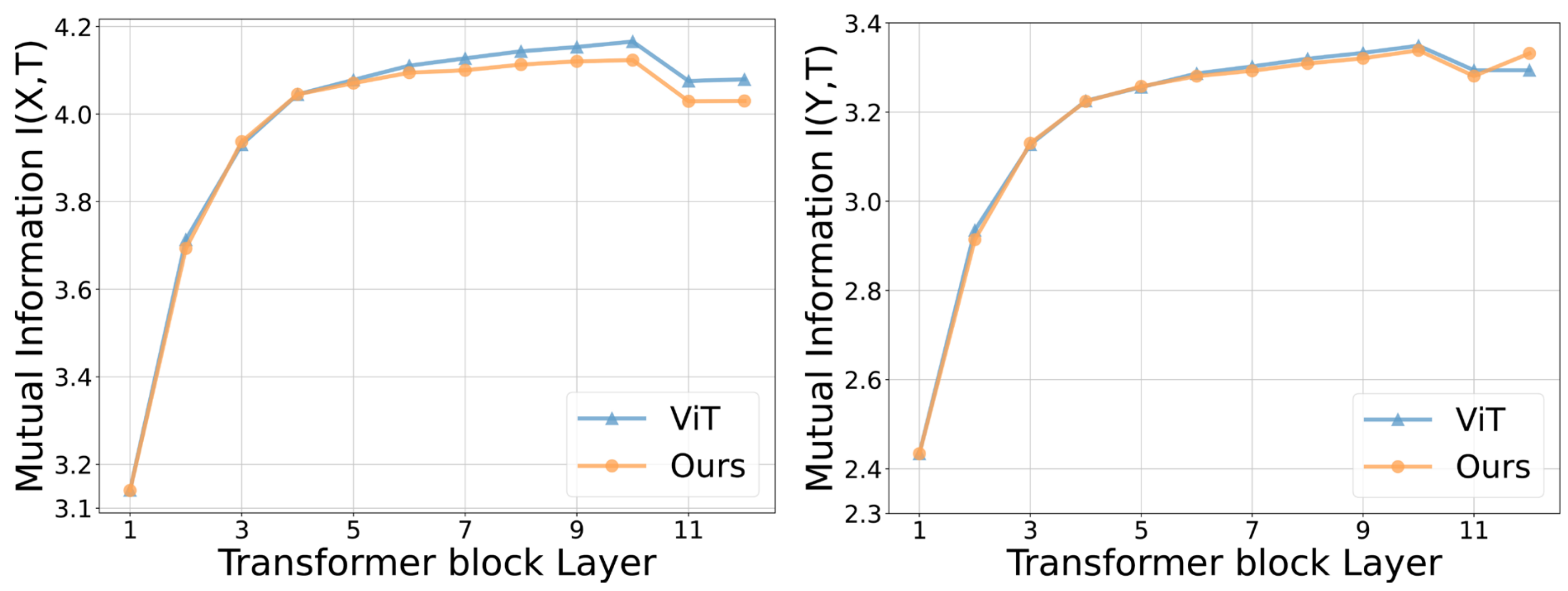}}
\subfigure[Stanford Dogs]{\includegraphics[height=1.4in,width=3.4in,angle=0]{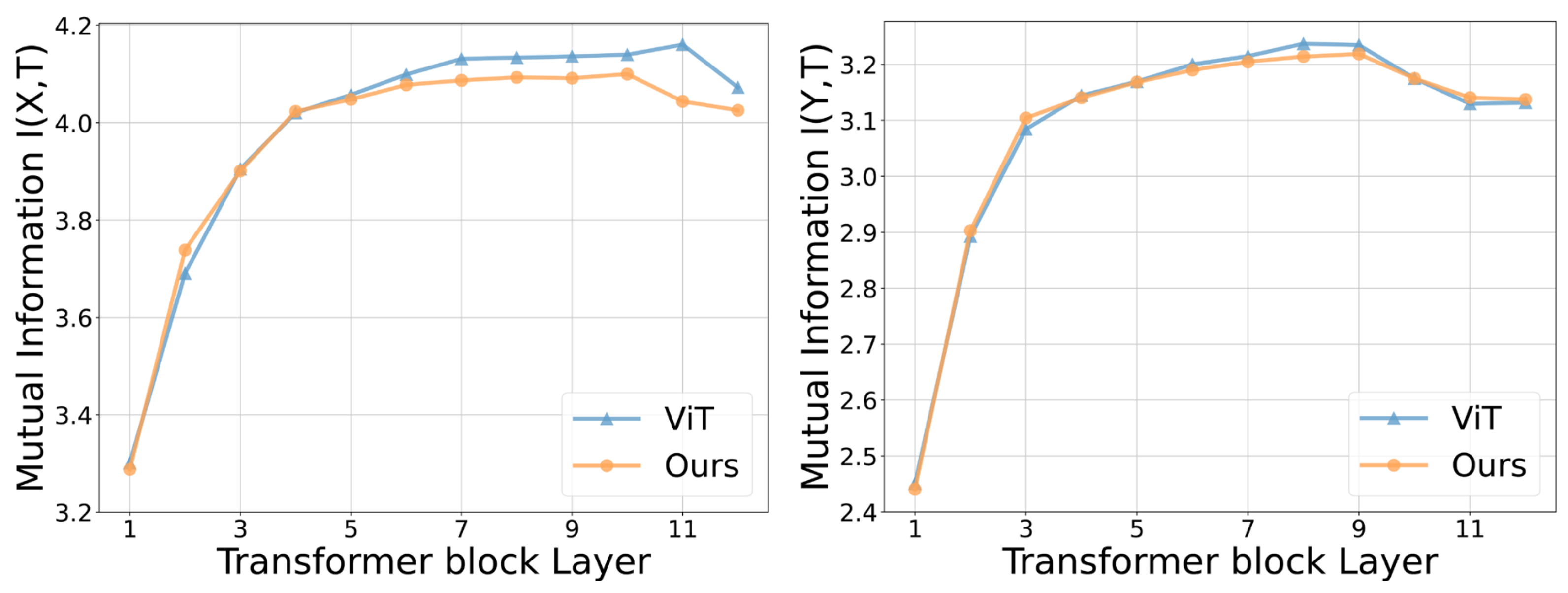}}
\subfigure[NABirds]{\includegraphics[height=1.4in,width=3.4in,angle=0]{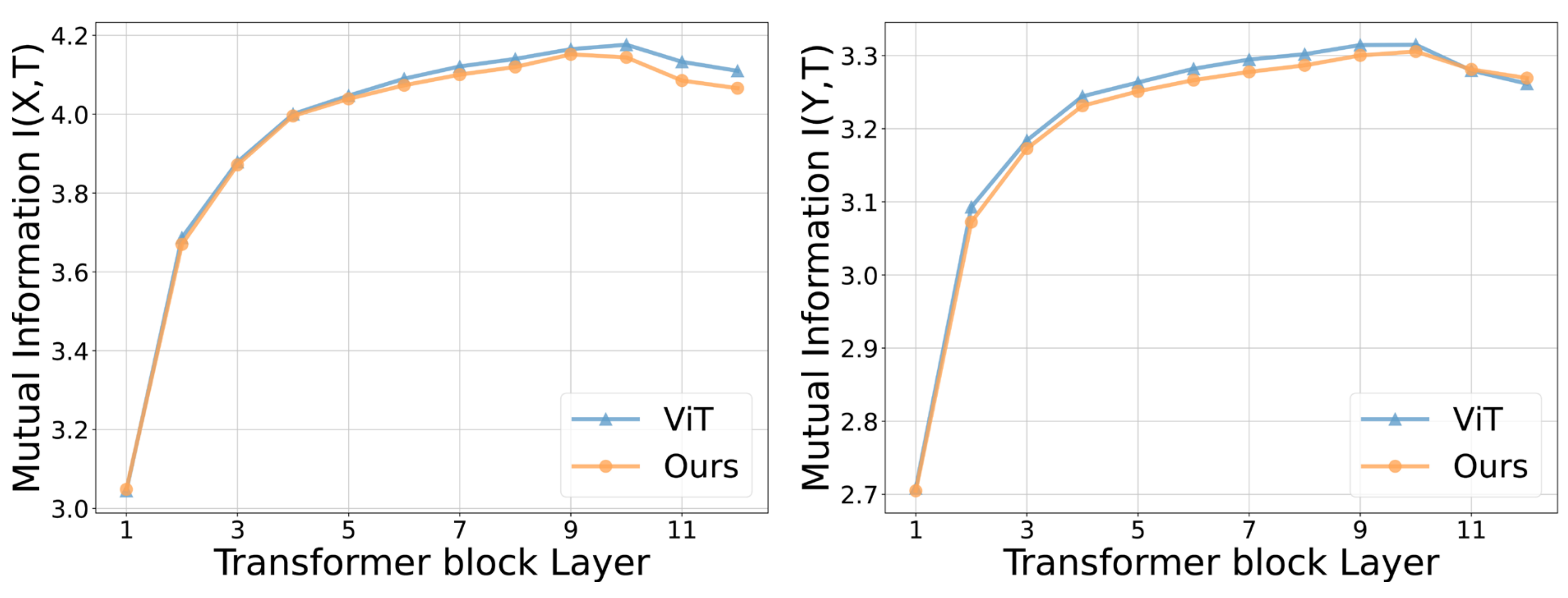}}
\setlength{\abovecaptionskip}{-0.1cm} 
\caption{Display of mutual information.The abscissa is the number of layer, and the ordinate is the mutual information. The left and right graphs show the result of class token with input $X$ , and the class token with label $Y$, respectively.}
\label{MI}
\end{figure}

%%%%%%%%%%%%%%%%%--table for Ablation Study--%%%%%%%%%%%%%%%%%%%%
\renewcommand{\arraystretch}{1.2}
\begin{table}[h]
\setlength{\tabcolsep}{8pt}
 \begin{center}
	\setlength{\abovecaptionskip}{-0.04cm}
\caption{Ablation studies of the $R^{2}$-Trans on the CUB-200-2011 dataset, the Stanford Dogs dataset, and the NABirds dataset.}
  \label{tab:tab5}
 \begin{tabular}{c|cccc}  
 \bottomrule 
  	\ Dataset  & Baseline  & Ours w/o D & Ours w/o I & Ours\\  
    \hline
            CUB      & 90.3  & 91.2 & 91.3 & 91.5  \\
            Dogs     & 92.0  & 92.4 & 92.5 & 92.8  \\
            Nabirds  & 89.6  & 90.0 & 89.8 & 90.2  \\
  \hline  
\end{tabular}
\end{center}
\vspace{-0.2cm} 
\end{table}

To evaluate the proposed $R^{2}$-Trans, we performed ablation experiments on all three datasets. Use CUB as an example, we first evaluate the performance of the basic model in FGVC tasks, we implement "baseline" by learning the fine-grained objectives without any action. 
Second, we evaluate the effectiveness of our proposed BDMM. To this end, we implement an "Ours w/o D" baseline by learning the fine-grained objectives without using the mask action. The experimental results are reported in Table~\ref{tab:tab5}, from which we can observe that without using BDMM, our approach receives a 0.3\% degeneration in terms of accuracy. 
Then, we evaluate the effectiveness of our proposed IB loss, which has never been used in the existing FGVC models. Specifically, we implement an "Ours w/o I" baseline by whether or not to add a constraint on the amount of information in $T$ to the loss function. The experimental results are reported in Table V. By comparing the performance between “OURS w/o I” and OURS, we can observe that the information bottleneck loss may impact the performance of our approach by 0.2\% in terms of accuracy. Similar results can be observed in two other data sets. This indicates that the information bottleneck loss can better guide the learning process to find the discriminant features and avoid the interference of redundant information.  

\begin{figure*}[t]
\centering
\subfigcapskip=5pt
\begin{tabular}{ccc}
\subfigure[CUB-200-2011]{\includegraphics[width=0.32\linewidth]{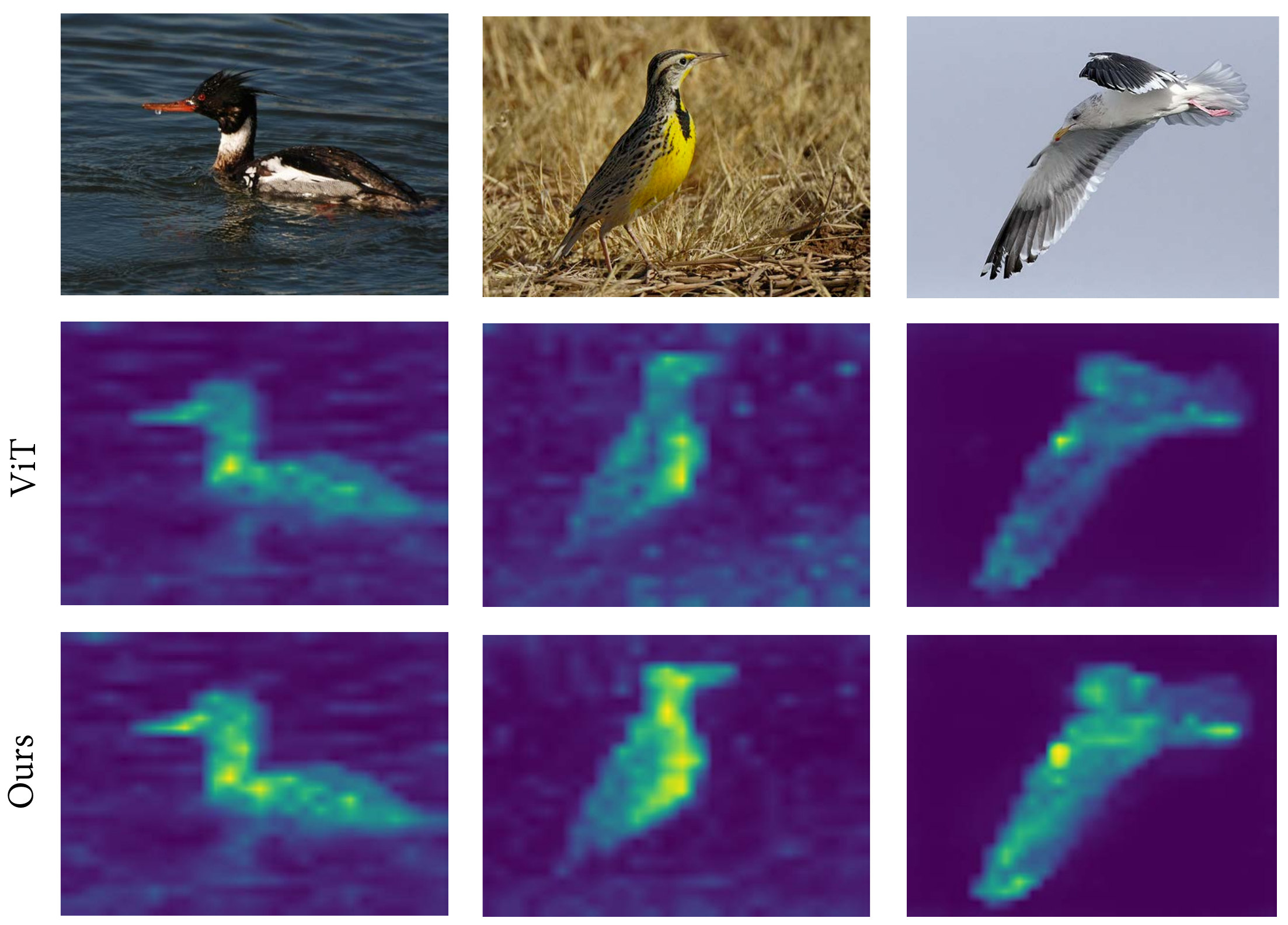}}\hspace{-2mm}
& 
\subfigure[Stanford Dogs]{\includegraphics[width=0.31 \linewidth]{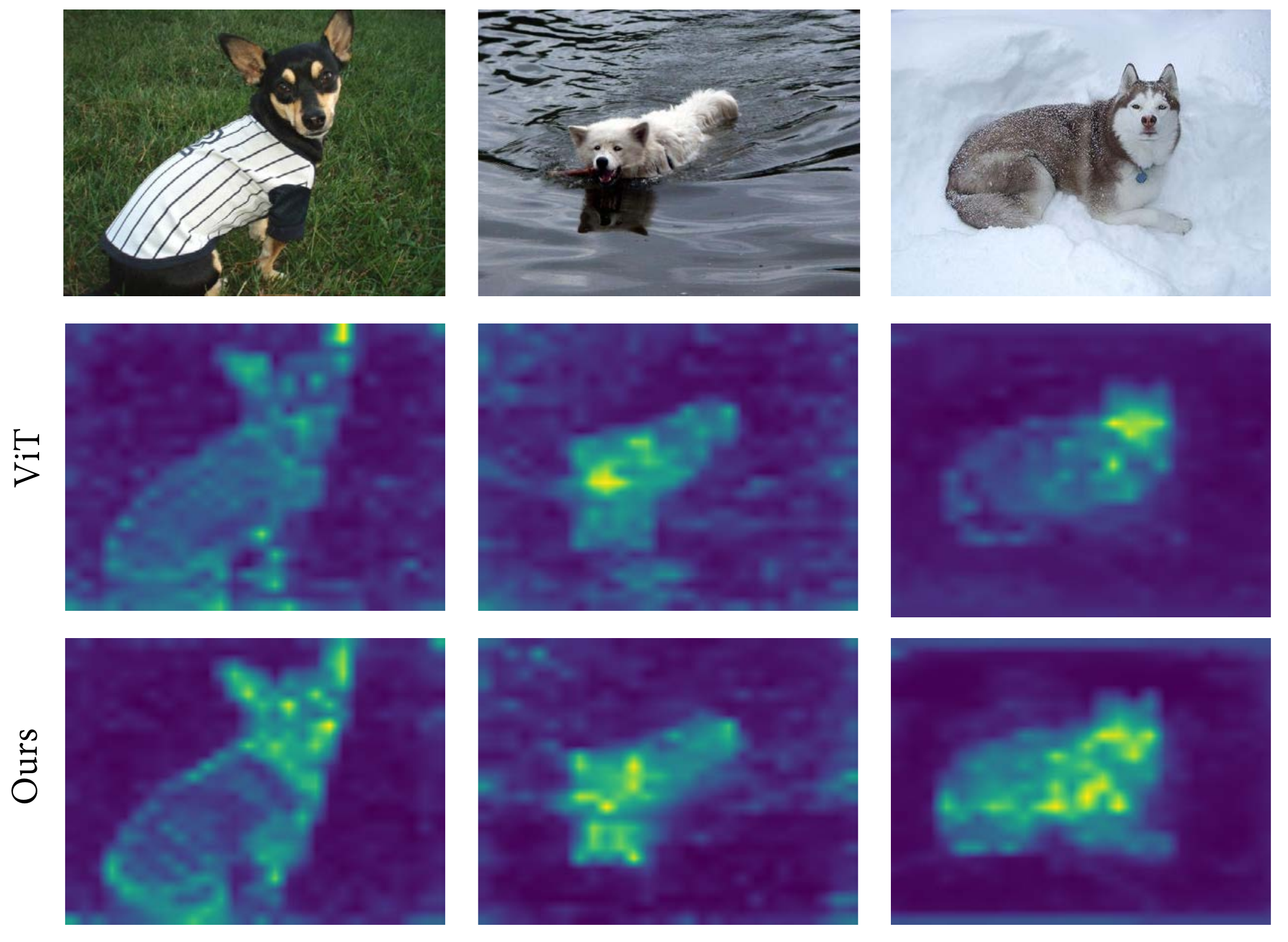}}\hspace{-2mm}
& 
\subfigure[NABirds]{\includegraphics[width=0.32 \linewidth]{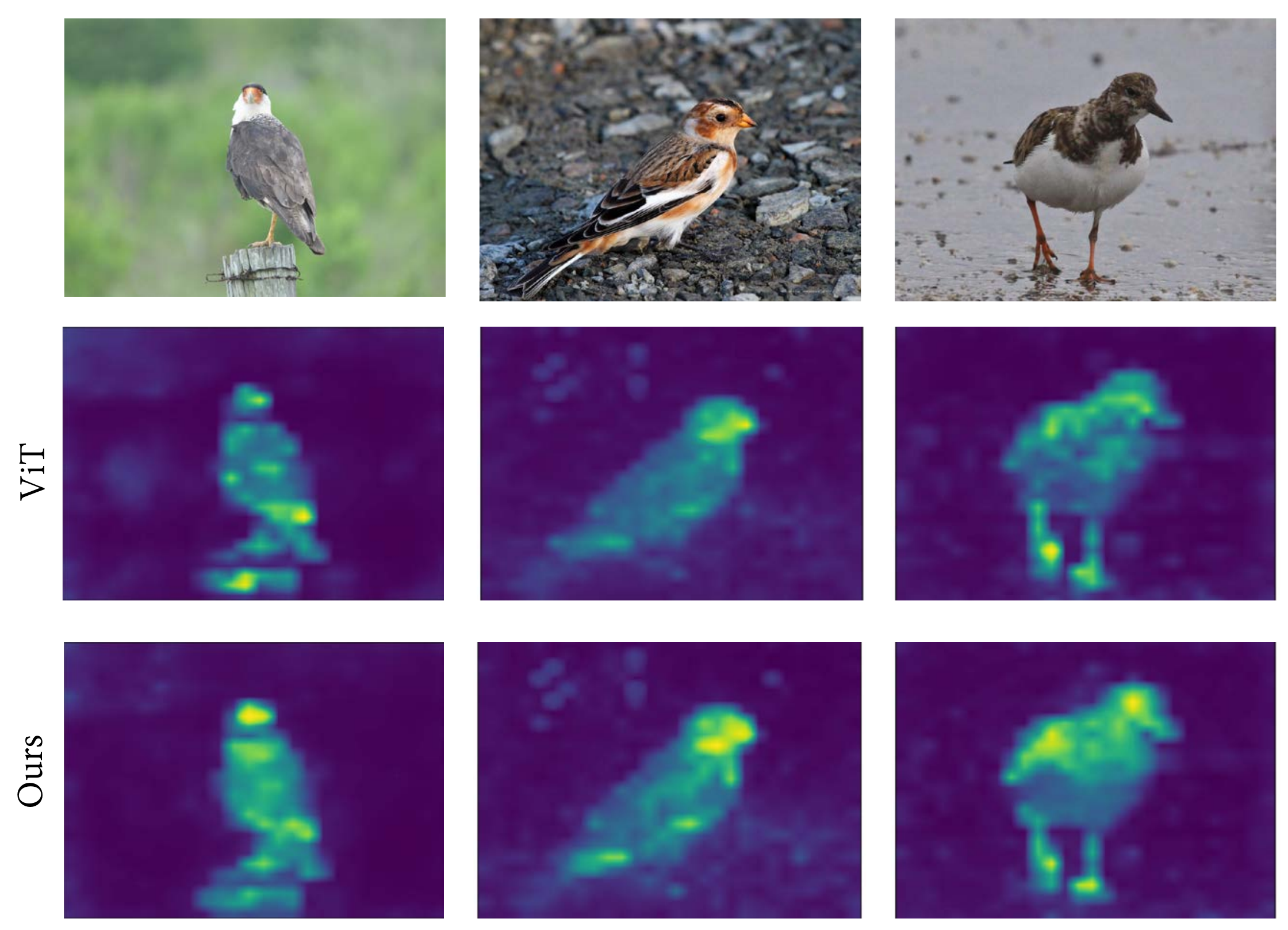}}
\end{tabular}
\setlength{\abovecaptionskip}{-0.1cm} 
\caption{Visualization analysis of CUB and Dogs datasets. The first line is the original image, and the second action is the attention result of the ViT. Our results are shown in the third line.}
\label{Attention visualization}
\end{figure*}

To further illustrate the effect of our method, we quantitative analysis the mutual information between each layer of class token and input image $X$ and label $Y$ in the model. The result is shown in Figure~\ref{MI}, where the blue line and the red line represent the results of the original model~((without IB regularization) and $R^{2}$-Trans respectively. As can be seen, our method compresses the information of $X$ and $T$ in the training process, but the information of $T$ and $Y$ remains the same. One possible reason is that ViT flattens the segmented image patches as patch tokens, then combines them with $T$ as input to Transformer Encoder. After coding, the $T$ results are extracted to predict categories. This architecture forces the dissemination of information between patch tokens and $T$. At the shallow layer, the interaction between patches and $T$ is limited. Therefore, the compression of redundant information cannot be fully manifested. The final results proves that our method is reliable and can effectively reduce redundant information, especially in deeper layers.

Figure~\ref{Attention visualization} visualizes the attention map of $R^{2}$-Trans on the target, which is used to qualitatively reveal the validity of our method. As can be seen, our model is more focused on the target instance after the redundancy information is compressed.

\section{Conclusion}
Locate the part area of the target has proved an important step for FGVC, most of the previous works mainly pay attention to training object part detectors to provide conditions for subsequent amplification of details. This implicitly ignores the use of partial discriminative information in environmental cues. In addition, the interaction of features introduces a great deal of redundancy. To avoid the problem, this paper proposes a novel $R^{2}$-Trans as a way of efficiently reducing redundancy features to improve FGCV. In particular, we observed that a large number of repetitive backgrounds may provide implicit information about the target, and the BDMM can adaptively and intelligently reduce the background of repeated information in the input space. Moreover, by leveraging information bottleneck theory, we also explore the feasibility of reducing target feature redundancy in feature space, which is verified to achieve promising performance and make our model was more interpretable. Compared with several SOTA methods on FGVC dataset, our $R^{2}$-Trans approach achieves competitive performance. In the future, we would consider extracting local mutual information of the target to make our model more intuitive and efficient.

\ifCLASSOPTIONcaptionsoff
  \newpage
\fi

\bibliographystyle{IEEEtran}
\bibliography{bib.bib}

\begin{IEEEbiography}[{\includegraphics[width=1in,height=1.25in,clip,keepaspectratio]{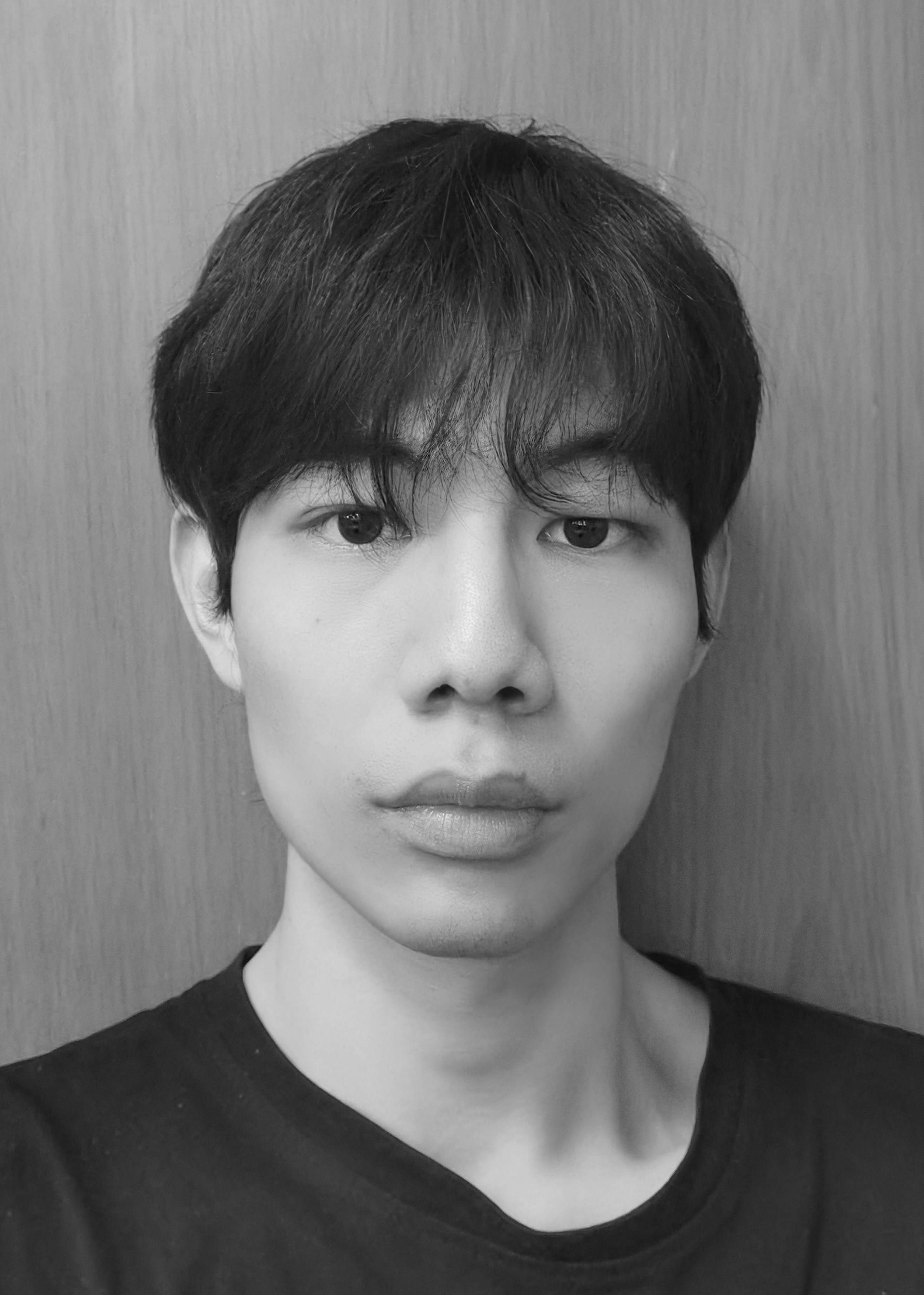}}]{Wang Yu}
received the B.S. degrees in School of Electronic Information and Communications, Huazhong University of Science and Technology in 2019. He is currently a full-time M.S student in the School of Electronic Information and Communications, Huazhong University of Sciences and Technology (HUST), China. His current research interests include machine earning and computer vision.
\end{IEEEbiography}

\begin{IEEEbiography}
[{\includegraphics[width=1in,height=1.25in,clip,keepaspectratio]{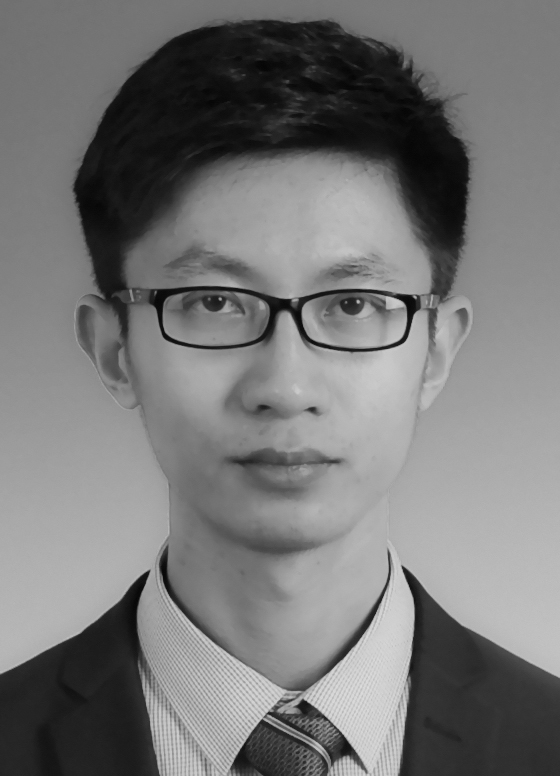}}]{Shuo Ye}
is currently a full-time Ph.D student in the School of Electronic Information and Communications, Huazhong University of Sciences and Technology (HUST), China. His current research interests span computer vision and voice signal processing with a series of topics, such as automatic speech recognition and fine-grained image categorization. 
\end{IEEEbiography}

\begin{IEEEbiography}
[{\includegraphics[width=1in,height=1.25in,clip,keepaspectratio]{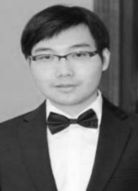}}]{Shujian Yu}
received his Ph.D. degree from the Department of Electrical and Computer Engineering at the University of Florida in 2019, with a Ph.D. minor in Statistics. He is now an associate professor at the Machine Learning Group in UiT – The Arctic University of Norway. His research interests lie on the intersection between machine learning and information theory. He received the 2020 International Neural Networks Society (INNS) Aharon Katzir Young Investigator Award for the contribution on the development of novel information theoretic measures for analysis and training of deep neural networks.
\end{IEEEbiography}
\vspace{-14cm} 

\begin{IEEEbiography}[{\includegraphics[width=1in,height=1.25in,clip,keepaspectratio]{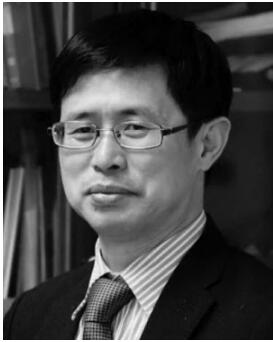}}]{Xinge You}
(SM’03) received the B.S. and M.S. degrees in mathematics from Hubei University, Wuhan, China, in 1990 and 2000, respectively, and the Ph.D. degree from the Department of Computer Science, Hong Kong Baptist University, Hong Kong, in 2004.
He is currently a Professor with the School of Electronic Information and Communications, Huazhong University of Science and Technology, Wuhan. His current research interests include image processing, wavelet analysis and its applications, pattern recognition, machine earning, and computer vision.
\end{IEEEbiography}

\end{document}